\documentclass[11pt,a4paper]{article}
\usepackage[hyperref]{emnlp-ijcnlp-2019}
\usepackage{times}
\usepackage{latexsym}

\usepackage{url}
\usepackage{graphicx} 
\usepackage{subfigure}
\usepackage{mathtools}
\usepackage{amsmath}
\usepackage{amssymb}
\usepackage{comment}
\usepackage{xcolor}
\usepackage{adjustbox}
\usepackage{lipsum}
\usepackage[draft]{todonotes}
\usepackage{booktabs}
\DeclareMathOperator*{\argmax}{\texttt{argmax}}
\DeclareMathOperator{\softmax}{\texttt{softmax}}
\DeclareMathOperator{\BiLSTM}{\texttt{BiLSTM}}
\DeclareMathOperator{\LSTM}{\texttt{LSTM}}
\DeclareMathOperator{\BiAtt}{\texttt{BiAtt}}

\def\bW{\mathbf{W}}
\def\bh{\mathbf{h}}
\def\bq{\mathbf{q}}
\aclfinalcopy 

\title{Zero-shot Text-to-SQL Learning with Auxiliary Task}

\author{
  Shuaichen Chang\footnotemark[2] \hspace{1mm} \thanks{\hspace{1mm}Work done during an internship at JD AI Research} \\chang.1692@osu.edu \\\And Pengfei Liu \footnotemark[4]\hspace{1mm}   \\pfliu14@fudan.edu.cn \\\And Yun Tang \footnotemark[3] \\yun.tang@jd.com \\\AND Jing Huang\footnotemark[3] \\jing.huang@jd.com \\\And Xiaodong He\footnotemark[3] \\xiaodong.he@jd.com\\\\
\footnotemark[2]\hspace{0.8mm}  Department of Computer Science and Engineering, The Ohio State University \\
\footnotemark[4]\hspace{0.8mm}School of Computer Science, Fudan University \\
\setcounter{footnote}{2}
\footnotemark[3]\hspace{0.8mm} JD.COM AI Research \\\And Bowen Zhou\footnotemark[3] \\bowen.zhou@jd.com
}

\date{}

\begin{document}
\maketitle
\begin{abstract}

Recent years have seen great success in the use of neural seq2seq models on the text-to-SQL task.
However,  little work has paid attention to how these models generalize to realistic unseen data, which naturally raises a question: does this impressive performance signify a perfect generalization model, or are there still some limitations?

In this paper, we first diagnose the bottleneck of text-to-SQL task by providing a new testbed, in which we observe that existing models present poor generalization ability on rarely-seen data.
The above analysis encourages us to design a simple but effective auxiliary task, which serves as a supportive model as well as a regularization term to the generation task to increase the model’s generalization.
Experimentally, We evaluate our models on a large text-to-SQL dataset WikiSQL.
Compared to a strong baseline coarse-to-fine model, our models improve over the baseline by more than 3\% absolute in accuracy on the whole dataset. More interestingly, on a zero-shot subset test of WikiSQL, our models achieve 5\% absolute accuracy gain over the baseline, clearly demonstrating its superior generalizability.

\end{abstract}

\section{Introduction}
\label{intro}

Text-to-SQL has recently attracted much attention as a sequence-to-sequence learning problem due to its practical usage for search and question answering~\cite{dong2016acl,zhong2017seq2sql,xu2017sqlnet,cai2018ijcai,yu2018naacl,dong2018acl,finegan-dollak2018acl,Yu2018emnlp,wang2018EG,Shi2018IncSQL}. The performance on some text-to-SQL tasks has been improved progressively \cite{dong2018acl,wang2018EG,Shi2018IncSQL} in recent years.
As pointed out in \cite{finegan-dollak2018acl}, when evaluating models on text-to-SQL tasks, we need to measure how well the models generalize to realistic unseen data, which is very common in the real applications. 

Most of the previous text-to-SQL tasks assumed that all questions came from a fixed database and hence share one global table schema. This assumption is useful for some specific applications such as booking flights \cite{dahl1994atis} and searching GEO \cite{zelle1996geo}, but not applicable to many real scenarios when different questions involve querying on different tables. \cite{zhong2017seq2sql} addressed this problem and generated a dataset called WikiSQL, which is by far the largest text-to-SQL benchmark dataset.


In WikiSQL many tables have different table schemas and each table has its own limited labeled data. 
One common approach is to encode the table column names in the input to the training of an encoder-decoder model \cite{yu2018naacl,dong2018acl}. 
\cite{yu2018naacl} proposed to utilize high-level type information to better understand rare entities and numbers in the natural language questions and encode these information from the input. These type information come from either external knowledge graph, a column or a number. This approach of TypeSQL \cite{yu2018naacl} was proven to be effective on WikiSQL when it is required for the model to generalize to new tables.


We observe that a text-to-SQL encoder-decoder model implicitly learn a mapping between entities in natural language questions to column names in tables. The model is likely to fail on mapping to new table column names that it never sees before. Hence if we learn a better mapping from question words to table column names, then the text-to-SQL generation model would be better generalized to new tables. With this in mind, we introduce an auxiliary model to enhance the existing generation model.

Specifically, we propose a novel auxiliary \textbf{mapping task} besides traditional text-to-SQL \textbf{generation task}. Here we explicitly model the mapping from natural language entities to table column names. The mapping model serves as an supportive model to the specific text-to-SQL task as well as regularization to the generation model to increase its generalization. These two tasks are trained together with a multi-task learning loss.
In practice, we adopt the coarse-to-fine decoder as the prototype of our generation model due to their superior performance in text-to-SQL tasks. And the generation model is further improved by introducing bi-attention layer (question-to-table attention and table-to-question attention) \cite{Seo2017iclr} and attentive pooling layer \cite{bowen2016arxv}.

We test our models on WikiSQL, with emphasis on a  \textsc{zero-shot} subset, where the table schemas of the test data never occur in the training data. Compared to the coarse-to-fine model, our models improve over the baselines by $3\%$ absolute in accuracy, achieve execution accuracy of $81.7\%$. In particular, on the \textsc{zero-shot} test part of WikiSQL, our models achieve even more gain, with $5\%$ improvement in accuracy over the baseline model.~\footnote{Our code will be released after paper is reviewed.}

In summary our contributions in this paper are three-fold:

    1) We find the existing testbed covers up the true generalization behavior of neural text-to-SQL models,
    and propose a new {\it zero-shot} test setting.
    
   2) We improve the generalization ability of existing models by introducing an auxiliary task, which can explicitly learn the mapping between entities in the question and column names in the table.
   
    3) The \textit{zero-shot} evaluation not only shows the superior performance of our proposed method compared with the strong baseline but makes it possible to explain where the major gain comes from.
    


\section{Background}

\subsection{Text-to-SQL Task}
\label{sec:background-task}
{\it Text-to-SQL} task can be formulated as a conditional text generation problem, in which a question $\mathcal{Q}$ and a table $\mathcal{C}$ are given, the goal is to generate a SQL language $\mathcal{Y}$.



 Figure \ref{figure:sql_sketch} illustrates WikiSQL output format which consists of three components: \texttt{AGG}, \texttt{SEL}, and \texttt{WHERE}. Particularly,  \texttt{WHERE} clause contains multiple conditions where each condition is a triplet with the form of (condition\_column,  condition\_operation, condition\_value).





\paragraph{Encoding Layer} The question $\mathcal{Q}$ and corresponding table schema $\mathcal{C}$ are first translated into the hidden representation by a BiLSTM sentence encoder:
\begin{align*}
    \mathbf{h}^{q}_t &=  \BiLSTM(\overrightarrow{\mathbf{h}}^{q}_{t-1},\overleftarrow{\mathbf{h}}^{q}_{t+1}, \mathbf{q}_t, \mathcal{\theta}) \\
    \mathbf{h}^{C}_t &=  \BiLSTM(\overrightarrow{\mathbf{h}}^{C}_{t-1},\overleftarrow{\mathbf{h}}^{C}_{t+1} \mathbf{C}_t,\mathcal{\theta})
\end{align*}
where $\mathbf{q}_t$ is embedding of question word $q_t$ and $\mathbf{C}_t$ is the representation of a column name $C_t$ which consists of words $c_t^1, \cdots, c_t^{|C_t|}$. The first and last hidden state of a BiLSTM over $C_t$ is concatenated as $\mathbf{C}_t$.



\paragraph{Decoding Layer}
Different from traditional text generation tasks, which share a decoder cross time-steps, in Text-to-SQL task, different decoders are designed in terms of different operations.
Generally, these decoders can be classified two types: \textsc{cls} for classifier, and \textsc{pt} for pointer.

\textsc{cls} is used for the operations, such as \texttt{AGG} and \texttt{COND\_OP}: 
\begin{align}
    \textsc{cls}(\mathbf{h}^{d}_t, \theta) &=  \softmax(\mathbf{h}^{d}_t, \theta)
\end{align}
where $\mathbf{h}^{d}_t$ is one decoder hidden representation.



\textsc{pt} can be used to choose a proper column or word from a set of column or words. Formally, We refer to $\mathbf{h}^d_t$ as a pointer-query vector and $K=\{k_1,..k_{|K|}\}$ as a set of pointer-key vectors, and predict the probability of choosing each key:
\begin{align}
    \textsc{pt}(\mathbf{h}^d_t,\mathbf{K}) &=\softmax (u)  \label{pointer-eq}
\end{align}
$u_i$ can be obtained as:
\begin{align}
    \mathbf{u}_i &=\mathbf{v}^T\tanh{(\bW[\mathbf{h}^d_t,k_i]+b)},  \quad i \in (1,...,|K|)
\end{align}

\begin{figure}[!t]
    \adjustbox{}{\textbf{SELECT} \texttt{\$AGG} \texttt{\$SEL}}
    \adjustbox{}{(\textbf{WHERE} \texttt{\$COND\_COL} \texttt{\$COND\_OP} \texttt{\$COND\_VAL})}
    \adjustbox{}{(\textbf{AND} \texttt{\$COND\_COL} \texttt{\$COND\_OP \texttt{\$COND\_VAL}})*}
    \caption{SQL Sketch. The tokens starting with ``\$" are slots to fill. ``*" indicates zero or more \textbf{AND} clauses.}
    \label{figure:sql_sketch}
\end{figure}

\begin{figure}[t]
\setlength{\belowcaptionskip}{-0.1cm}
\centering
\includegraphics[width=0.8\linewidth]{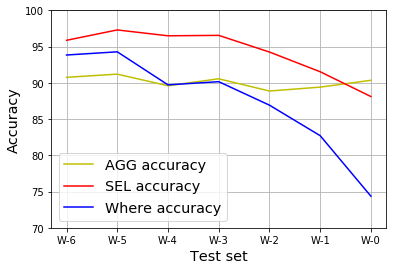}
\caption{ Break down accuracy of a strong baseline model \cite{dong2018acl}. X-axis represents different subsets of WikiSQL test set, split by how many times a table occurs in training data. Splitting details are in Table \ref{data-size}.
}\label{fig:intro}
\end{figure}

\subsection{Diagnosing the Bottleneck of Text-to-SQL }

The existing testbed covers up the true generalization behavior of existing models. To address this problem, we provide a new testbed by breaking down the testing samples.
Specifically, we analyze the generalization problem on table aware {\it Text-to-SQL} tasks, by testing previous state-of-the-art model \cite{dong2018acl} on different tables which occur different times in training set. We observe the following problems 
based on Figure \ref{fig:intro}:
\begin{itemize}
\item \texttt{WHERE} clause performance is more sensitive to how many times the table has been seen in the training data;
\item The performance of \texttt{WHERE} would get a big drop once the table in test set is not present in the training data, i.e. zero-shot testing case.
\end{itemize}

Despite of the importance of the generalization problem of unseen tables, few work explored it due to the lack of appropriate datasets.
The WikiSQL dataset was originally constructed to ensure that the training and test set have disjoint set of tables, which can provide a test bed for generalization test of new tables. However, we find that the current version of WikiSQL test cannot guarantee this because different tables extracted from different wiki pages may share the same table schema (i.e. table column names), even though their table content may not be the same.

The above problems motivate us to explicitly model the mapping between words in question and table column names, and test the model generalization to new tables on the true zero-shot sub testset of WikiSQL.
\section{Model}

Our model consists of a seq2seq model for the SQL generation task (largely following the baseline coarse-to-fine model), and a mapping model as a auxiliary task to explicitly map question words to table schema (column names).

\subsection{Main Generation Model}



\paragraph{\texttt{Encoder}} we follow section \ref{sec:background-task} to obtain question and schema hidden representation $\mathcal{H}^q$ and $\mathcal{H}^c$. To enhance the interaction between question words $q$ and column name $c$, a bi-attention is used to generate final question and table schema representation:
\begin{align*}
    \bar{\mathcal{H}}^q, \bar{\mathcal{H}}^c = \BiAtt(\mathcal{H}^q,\mathcal{H}^c,\theta)
\end{align*}





 Considering the nature of structured SQL, we follow previous works 
 to use different sub-decoders for \texttt{AGG}, \texttt{SEL} and \texttt{WHERE} clause. Especially, our \texttt{WHERE} decoder is adapted from the baseline model \cite{dong2018acl}.

\paragraph{\texttt{AGG} and \texttt{SEL} Decoder}
Each SQL only contains one \texttt{AGG} and \texttt{SEL}, so we generate \texttt{AGG} and \texttt{SEL} based on entire question representation. Since different words or phrases in question do not equally contribute to the decisions of  \texttt{AGG} and \texttt{SEL}, we employ an attentive pooling layer over $\bar{\mathcal{H}}^q$ to generate final hidden representation ${\bq}^{SEL}$ for \texttt{AGG} and \texttt{SEL}.



We feed ${\bq}^{SEL}$ into \textsc{cls} layer generate the aggregation operation \texttt{AGG} and 
 meassure the similarity score between ${\bq}^{SEL}$ and each column name $\bar{\mathbf{C}}_j$ to predict \texttt{SEL} by \textsc{Pt} layer in (\ref{pointer-eq}):
\begin{align*}
    y^{AGG}  & = \textsc{cls}({\bq}^{SEL}, \theta) \\
    y^{SEL}  &= \textsc{pt}(\bq^{SEL},\bar{\mathcal{H}}^{c})
\end{align*}

\begin{figure*}[t]
\centering
\begin{subfigure}
    \centering
    \includegraphics[width=0.7\textwidth]{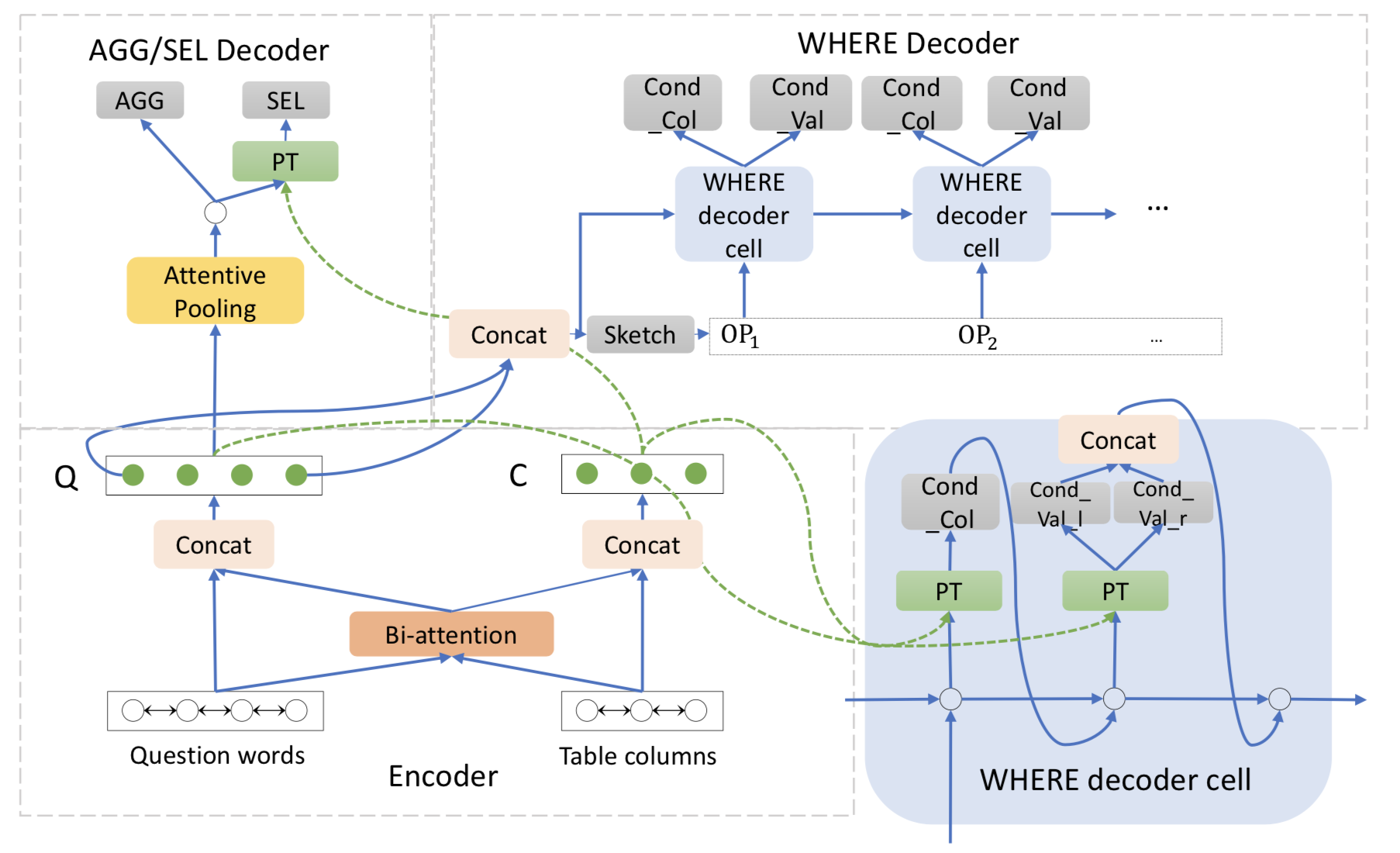}
    \label{figure:-G-model}
\end{subfigure}
\begin{subfigure}
    \centering
    \includegraphics[width=0.7\textwidth]{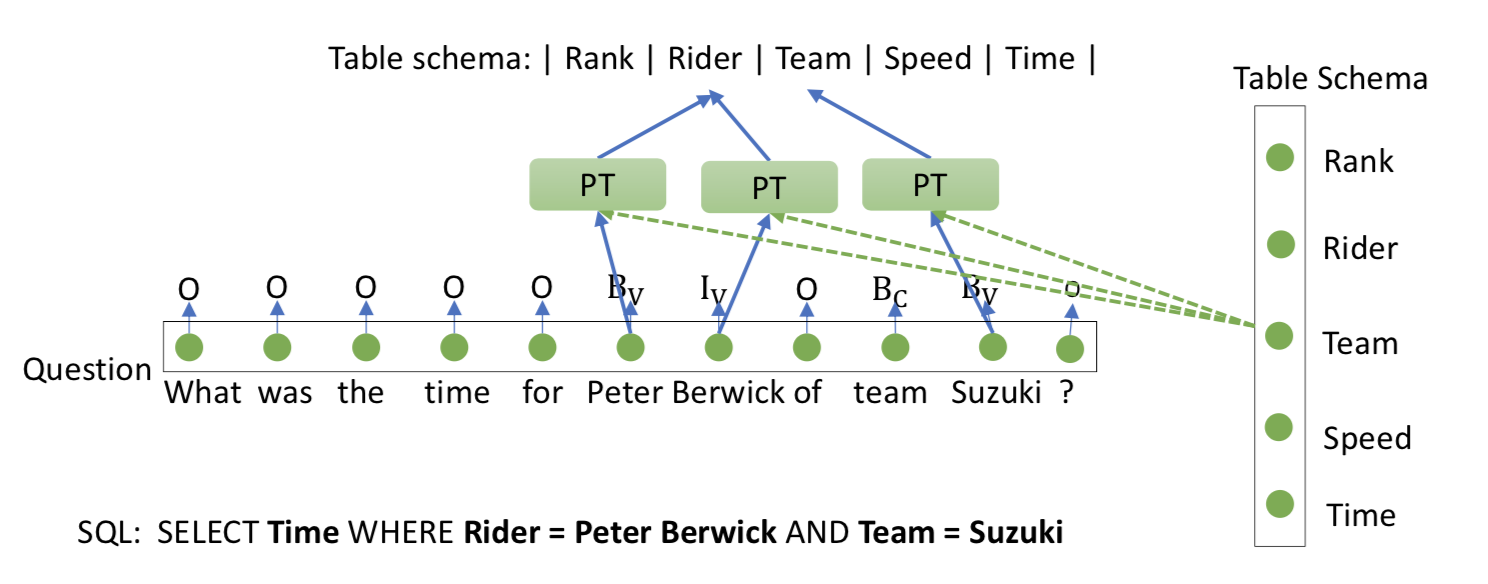}
    \label{figure:M-model2}
\end{subfigure}
\caption{Illustration of our model. The upper figure is the text-to-SQL generation model which consists of three parts: encoder (lower left), AGG/SEL decoder (upper left) and where decoder (upper right). Lower right is \texttt{WHERE} decoder cell. 
The bottom figure is our auxiliary mapping model with the ground-truth label of an example. Question word is mapped to a column only when it is tagged as part of a condition value ($\rm B_v$ or $\rm I_v$).}
\label{figure:model}
\end{figure*}



\paragraph{\texttt{WHERE Decoder}} We took the \texttt{WHERE} decoder from the-state-of-the-art model \cite{dong2018acl}, which first generates a slot sketch of \texttt{WHERE} clause and transform the SQL generation into a slot filling problem. There are 35-category \texttt{WHERE} clauses in WikiSQL and each one is subsequence of \texttt{WHERE} clause which skip the \texttt{COND\_COL} and \texttt{COND\_VAL}. For example, "\texttt{WHERE}  $=$  \texttt{AND} $>$ " is a sketch of \texttt{WHERE} clause which has 2 conditions. We first predict the sketch $\alpha$ based on $\bar{\mathcal{H}}^q$:
\begin{align*}
y^{\alpha}  & = \textsc{cls}({\bq}^{Where}, \theta) ,
\end{align*}
where ${\bq}^{Where} = [\bar{\mathbf{h}}_1^q,\bar{\bh}^q_{|\mathcal{Q}|}]$.


Once $y^{\alpha}$ is predicted, we obtain the \texttt{COND\_OP} sequence it represents. We embed each operation in \texttt{COND\_OP} sequence and feed them into \texttt{WHERE}-decoder cell. As Figure~\ref{figure:model} shows, the \texttt{WHERE}-decoder cell takes one \texttt{COND\_OP} as input and output \texttt{COND\_COL} and \texttt{COND\_VAL} for each decoder time step, while each decoder time step has 3 LSTM time steps. For $i$th condition, $x^d_{i,1}$,$x^d_{i,2}$,$x^d_{i,3}$ are $\texttt{COND\_OP}_i$ and $\texttt{COND\_COL}_i$ and $\texttt{COND\_VAL}_i$ and output $y^d_{i,1}$,$y^d_{i,2}$ are probability distribution of the index of $\texttt{COND\_COL}_i$ and the span of $\texttt{COND\_VAL}_i$. We do not have output for each $y^d_{i,3}$ because the input of next time step is given by pre-predicted $\texttt{COND\_OP}_{i+1}$.
The lstm-cell is updated 3 times for each decoder time step:
\begin{small}
\begin{equation*}
h^d_{i,j}=\begin{cases}
\LSTM (x^d_{i,j},h^d_{i,j-1})\quad & j\neq 1\\
\LSTM (x^d_{i,j},h^d_{i-1,3}) \quad & j=1
\end{cases}
\end{equation*}
\end{small}

The output layers for \texttt{COND\_COL} and \texttt{COND\_VAL} are both pointer layer which are pointed to column names and question words to predict \texttt{COND\_COL} index and the left and right end $VAL^l$, $VAL^r$ of the span of \texttt{COND\_VAL} in question:
\begin{small}
    \begin{align*}
        y^d_{i,1}&= \textsc{PT}(h^d_{i,1},\mathcal{H}^c) \\
        y^d_{i,2}&=P(VAL_i^l|\cdot)\cdot P(VAL_i^r|VAL_i^l,\cdot)\\
        P(VAL_i^l|\cdot) &= \textsc{PT}(h_{i,2},\mathcal{H}^q)\\
        P(VAL_i^r|VAL_i^l,\cdot) &= \textsc{PT}([h_{i,2};\bar{\bh}_{VAL_i^l}^{q}],\mathcal{H}^q)
    \end{align*}
\end{small}

\subsection{Auxiliary Mapping Model} 
For a SQL query, each condition consists of three parts, \texttt{COND\_COL}, \texttt{COND\_OP} and  \texttt{COND\_VAL}.
Our mapping model aims to discover the mapping between condition column and condition value. The mapping prediction is based on question and table schema representation $\mathcal{H}^q$ and $\mathcal{H}^c$, which are shared with generation model. Optimization based on mapping task loss can improve the question and table schema representation.
An intuitive way to achieve mapping is to directly learn a mapping function from each word in question to column names.
However, since not all words in a question are condition values, it's challenging to take all words into consideration.
To address this problem, we propose a two-step mapping model, in which we first learn a detector to screen out condition values, and then we learn a mapping function from condition values to column names.



\paragraph{Condition Value Detection}
 Because the condition value sometimes covers multiple words, we label the span for condition values in questions with typical $\rm BIO$ \cite{nadeau2007survey} tags. We notice sometimes condition column names appear exactly in question, so the span of column name in question is also labeled with tags $\rm B_{c}$, $\rm I_{c}$ during training  when a column name appear in question. Altogether we have five tags $\rm B_{c}$, $\rm I_{c}$, $\rm B_{v}$, $\rm I_{v}$, $\rm O$, which represent the first word of condition column, subsequent word of condition column, the first word of condition value, subsequent word of condition value and outside, respectively. Figure~\ref{figure:model} illustrates our mapping model by giving the ground-truth label for an example. 

The mapping model takes encoding vector of question words $\bar{\mathcal{H}}^{q}= \bar{\bh}^{q}_1,..., \bar{\bh}^{q}_{|\mathcal{Q}|}$ and column names $\bar{\mathcal{H}}^{c}= \bar{\bh}^{c}_1,..., \bar{\bh}^{c}_{|\mathcal{C}|}$ as input. Mapping model first predict gate  $y^{tag}$:
\begin{align*}
    y^{tag}_i &=\argmax(\mathbf{v}_{tag}\tanh(\mathbf{W}_{tag} \bar{\bh}^{q}_i+\mathbf{b}_{tag})),
\end{align*}
where $\mathbf{W}_{tag} \in \mathbb{R}^{5*H} $ and $\mathbf{b}_{tag}\in \mathbb{R}^5$ are tagging parameters.

\paragraph{Value-column Mapping}
We only learn the mapping  for question words which are tagged as $\rm B_{v}$, $\rm I_{v}$:
\begin{align*}
    y_i^{map} &= \textsc{PT}(\bar{\bh}^{q}_i,\bar{\mathcal{H}}^{c}),  \ \ \  y_i^{tag} \in \{B_v, I_v\}
\end{align*}

\subsection{Loss Function}

We refer to the following ${L}_{gen}$ as generation task loss and $\mathcal{L}_{map}$ as mapping task loss.
\begin{align*}
    {L}_{gen} & = -  \sum_{i=1}^{|\mathcal{Y}|}  y_i^{op} \log(\hat{y}_i^{{op}}), \\
    {L}_{map} &=  -  \sum_{i=1}^{|\mathcal{Q}|}  y_i^{tag} \log(\hat{y}_i^{{tag}}) - \sum_{i=1}^{K}  y_i^{map} \log(\hat{y}_i^{{map}}),
\end{align*}
where $op$ represents different operations during decoder phase. $y$ and $\hat{y}$ denote the probability distribution of real label and predicted probability distribution. $K$ represents how many times words in question have been predicted as condition values. 

Finally, the overall loss can be written as:
\begin{align*}
    \mathcal{L} & = \sum_{i=1}^{N} \lambda L_{gen} + (1-\lambda) L_{map}   
\end{align*}
where $N$ is the number of training samples and $\lambda$ is hyper-parameter.

\section{Experimental Setup}


\subsection{Dataset}
    \label{sec:Dataset}

WikiSQL has over 20K tables and 80K questions corresponding to these tables. This dataset was designed for translating natural language questions to SQL queries using the corresponding table columns without access to the table content. This dataset is further split into training and testing sets that are separately obtained from different Wiki pages, assuming there is no overlap of tables between training and testing sets. However, we find in this split, $70\%$ question-table pairs in test set have the same table schema as those in the training set. This is because even train and test tables were obtained from different Wiki pages, these tables could still have the same table schema. For example, different football teams have their own Wiki page but each one have a table with the same schema recording match information. 

We split the test set based on the number of shots (the number of a table occurrences in training data). We report experiments on the original full WikiSQL test set as well as different subset based on the number of shots, especially on the {\it zero-shot} testing case. Statistics of new test sets  are in table \ref{data-size}.


    
    

\begin{table*}[!t]
    \begin{center}
    \begin{small}
        \begin{tabular}{p{4cm} c c c c c}
        \toprule
        \bf Model &\bf $ACC_{qm}$ & \bf $ACC_{ex}$ &\bf $ACC_{agg}$ & \bf $ACC_{sel}$ & \bf $ACC_{where}$ \\
		\midrule
		SEQ2SQL \cite{zhong2017seq2sql}  &- & 59.4\% & 90.1\% & 88.9\% & 60.2\% \\
		\midrule
		SQLNET \cite{xu2017sqlnet}  & 61.3\% & 78.0\% & 90.3\% & 90.9\% & 71.9\% \\
		\midrule
		TypeSQL \cite{yu2018naacl}   & 66.7\% & 73.5\% &  90.5\% & 92.2\% & 77.8\% \\
		\midrule
		COARSE2FINE \cite{dong2018acl}  & 71.7\% & 78.5\% & 90.4\% & 92.4\% & 84.2\% \\
		\midrule
		\midrule
		Gen-model w/o AP &  72.8\% &  79.4\%  & 90.2\% & 93.0\% & 84.7\% \\
		\midrule
		Gen-model & 73.5\% &  80.1\% & 90.3\% & 94.2\% & 84.8\% \\
		\midrule
		Full-model & \bf 75.0\% & \bf 81.7\%  &   90.5\% &  \bf 94.5\% & \bf 86.7\% \\
		\bottomrule
        \end{tabular}
    \end{small}
    \end{center}
    
    \caption{Overall and break down results on full WikiSQL dataset. $\rm ACC_{qm}$, $\rm ACC_{ex}$ are accuracy numbers of query match (ignore the order of conditions) and execution result, and $\rm ACC_{agg}$, $\rm ACC_{sel}$, $\rm ACC_{where}$ are the accuracy of \texttt{AGG}, \texttt{SEL}, \texttt{WHERE} clauses. The upper part are baseline models, and the lower part are our generation model Gen-model and the whole model Full-model which is the Gen-model with the auxiliary mapping model. Gen-model w/o AP is the generation model without attentive pooling. }
    \label{result-full}
\end{table*}

\begin{table} [!ht]
\begin{small}
    \begin{center}
        \begin{tabular}{lll}
        \toprule
         \bf dataset &  \bf number of shots & \bf \#questions \\
		\midrule
		W-full & [0,2045] & 15878  \\
		\midrule
		W-0 & 0 & 5201  \\
		\midrule
		W-1 & [1,5] & 1700 \\
		\midrule
		W-2 & [6,15] & 1842 \\
		\midrule
		W-3 & [16,40] & 1971 \\
		\midrule
		W-4 & [41,100] & 1654 \\
		\midrule
		W-5 & [101,500] & 1887 \\
		\midrule
		W-6 & [501,2045] & 1623 \\
		\bottomrule
        \end{tabular}
    \end{center}
    \caption{Statisitics of WikiSQL test set. W-full is original WikiSQL test set and W-0, W-1,$\cdots$, W-6 are subsets split by the number of shots (number of a table occurrences in the training data).}
    \label{data-size}
\end{small}
\end{table}

\subsection{Evaluation}
We follow the evaluation metrics in \cite{xu2017sqlnet} to measure the query synthesis accuracy: query-match accuracy ($\rm ACC_{qm}$) which measures the decoded query match the ground truth query without considering the order of conditions and execution accuracy ($\rm ACC_{ex}$) which measures the results from executing predicted queries. The accuracies are further break down into three categories: \texttt{AGG}, \texttt{SEL} and \texttt{WHERE}, as in 
\cite{xu2017sqlnet}.

\subsection{Model Configuration}
We use 300-dim Glove word embedding as our pre-trained embedding. Hidden size for all LSTM is 250 and hidden size in attention function is set to 64. The loss weight $\lambda$ is set to 0.5. A 0.5-rate dropout layer is used before each output layer. Each concatenation is followed by one full-connected layer to reduce the dimension to the original hidden or attention size. Test model is selected by the best performing model on validation set.

\section{Results and Analysis}

Table \ref{result-full} shows the overall and breakdown results on full WikiSQL dataset. We compare our models with strong baseline models on the original WikiSQL test data. All these models have no access to table content following \cite{zhong2017seq2sql}.

First our Gen-model with enhanced encoder/decoder improves over the baseline coarse-to-fine model by 1.6\% in accuracy of both $ACC_{qm}$ and $ACC_{ex}$.
Our Gen-model mainly improves on $ACC_{SEL}$ compared to baseline models. Ablation test shows the improvement is attributed to the attentive pooling in \texttt{SEL} decoding. 

Second our Full-model outperforms our single generation model by 1.5\% and 1.6\% in query-match accuracy and execution accuracy, achieving a very competitive new execution accuracy of 81.7\%. Break down results show Full-model mainly improves the accuracy over Gen-model on the \texttt{WHERE} clause, with 1.9\% accuracy gain.



\begin{figure}[!t]
  \centering
  \subfigure[\textsc{Where Decoder}]{
    \label{fig:arc1}
    \includegraphics[width=0.46\textwidth]{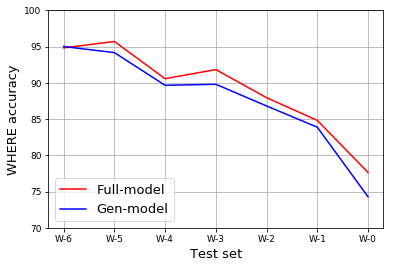} 
  }
  \subfigure[\textsc{AGG}/\textsc{SEL Decoder}]{
    \label{fig:arc2}
    \includegraphics[width=0.46\textwidth]{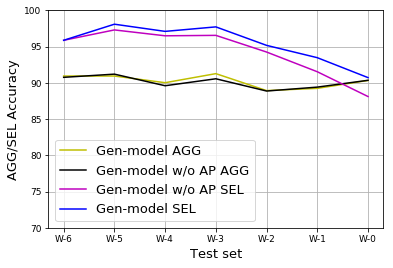} 
  }
 \label{result_with_testsets}
 \caption{Accuracy of Full-model and Gen-model in different test subsets. W-0 represents zero-shot setting. The frequency of the table has been seen in the training data decrease from W-6 to W-0.} 
\end{figure}

\begin{figure*}[!t]
    \centering
  \subfigure[Results on Unseen tables (W-0).]{
    \label{fig:left}
    \includegraphics[width=0.55\linewidth]{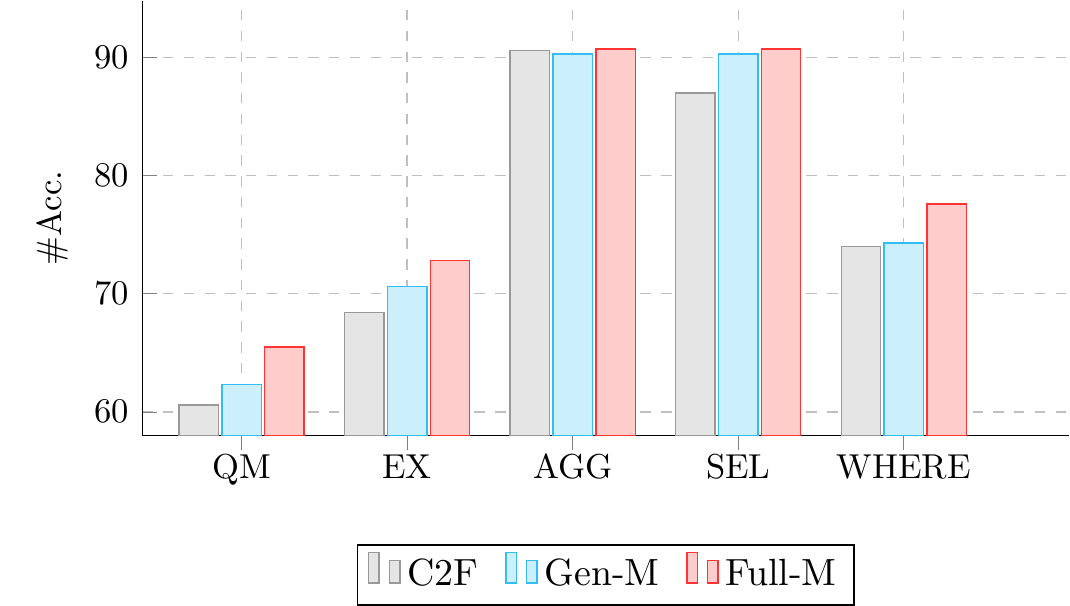} 
    }
  \subfigure[Results on seen/unseen columns.]{
    \label{fig:right}
    \includegraphics[width=0.28\linewidth]{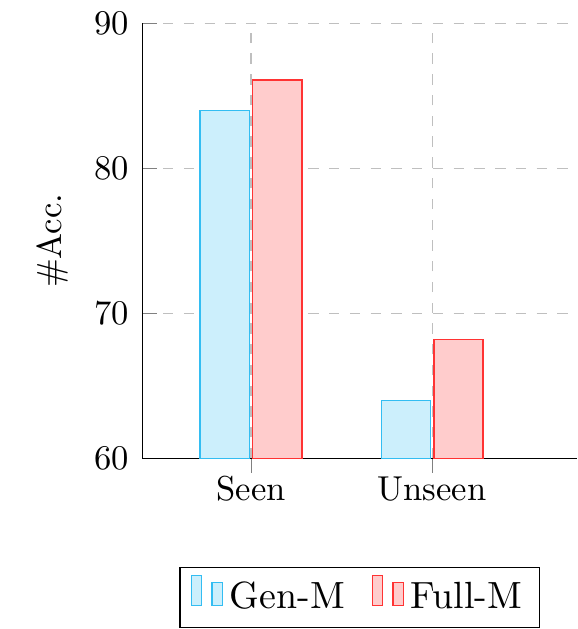}
    }
    \caption{\textit{C2F}, \textit{Gen-M} and \textit{Full-M} represent the baseline C2F model, and our proposed Gen-Model and Full-model respectively.}
\end{figure*}

\subsection{Training data amount}
Figure \ref{fig:arc1} illustrates Gen-model and Full-model accuracy of \texttt{WHERE} clause prediction on different test subsets from Table \ref{data-size}. Full-model is consistently better than single Gen-model in \texttt{WHERE} clause prediction. The biggest gap between Full-model and Gen-model in \texttt{WHERE} clause accuracy is on test subset W-0. This shows that Full-model  generalizes better than Gen-model for the unseen test tables. We also found that Full-model accuracy on W-4 is slightly lower than that on W-3. We believe this is due to the fact that table itself is the other fact affecting models' performance, in addition to the amount of training tables. 

Figure \ref{fig:arc2} again illustrates Gen-model outperforms Gen-model without attentive pooling on different amount of training data. 

\subsection{Zero-shot Test}
Figure \ref{fig:left} illustrates the results on {\it zero-shot} test case (i.e. W-0). Our Full-model outperforms baseline coarse-to-fine model by 4.9\% and 4.4\% in $\rm ACC_{qm}$ and $\rm ACC_{ex}$. The accuracy improvement over the baseline coarse-to-fine model lie on the \texttt{SEL} and \texttt{WHERE} clause, with 3.6\% accuracy gain on \texttt{WHERE} clause over the baseline.

Figure \ref{fig:arc1} shows \texttt{WHERE} clause accuracy has a big drop on zero-shot setting (W-0) compared to few-shot setting (W-1). We further analyze the reason of this degradation by looking into how the performance is affected by whether a column name is present in the training data. On unseen test table schema, 28\% column names never appear in training set, which makes question related to these columns harder. We further divide conditions in \texttt{WHERE} clauses into two classes, one class with condition column appearing in training, the other with condition column not appearing in training. We measure the accuracy of each class on the \texttt{WHERE} clause. The result is reported in Figure \ref{fig:right}. Full-model outperforms single generation model  by 4.2\% on unseen column names and 2.1\% on seen column names. This further shows the generalization ability of the Full-model.

\subsection{Case Study on Zero-shot Setting}

We also analyze the Full-model behavior on zero-shot test compared to the Gen-model alone. We first randomly sample 100 examples of which Full-model predicts correct on \texttt{WHERE} clause (Case-Correct in Table~\ref{tab:error-cases}), while Gen-model fails. We label the failure reasons of Gen-model into four categories (one example can belong to more than one categories): (a) wrong \texttt{COND\_COL} prediction, (b) wrong \texttt{COND\_VAL} prediction, (c) predicting extra conditions  or missing conditions and (d) others. Table~\ref{tab:error-cases} shows the majority of \texttt{WHERE} clause errors are in (a): wrong COND column name errors. 
We then randomly sample another set of 100 examples (Case-Wrong in Table~\ref{tab:error-cases}): Gen-model predicts \texttt{WHERE} clause correctly on these examples but Full-model fails. 
Table~\ref{tab:error-cases} indicates Full-model corrects  Gen-model mainly on wrong \texttt{COND\_COL} prediction, which shows our mapping task improves column name predictionin the generation task.

\begin{table}[]
    \begin{small}
    \centering
    \begin{tabular}{l c c c c}
    \toprule 
    \bf Examples   & \bf (a) & \bf (b) & \bf (c) & \bf others \\
    \midrule
     Case-Wrong & 63   & 22 & 18 & 4 \\
     \midrule
     Case-Correct & 71   & 19 & 10 & 3\\
    \bottomrule         
    \end{tabular}
    \caption{Number of samples in each error categories. }
    \label{tab:error-cases}
    \end{small}
\end{table}





\section{Related Work}
\label{relWork}


Recently neural network based approaches, especially sequence-to-sequence models have been applied to text-to-SQL successfully with progressively improving results
\cite{wang2017synthesizing,neelakantan2017iclr,iyer2017acl,yin2017acl,huang2018naacl,zhong2017seq2sql,xu2017sqlnet,cai2018ijcai,yu2018naacl,dong2018acl,finegan-dollak2018acl}.

Sketch-based approach is very effective, especially on WikiSQL task \cite{zhong2017seq2sql,xu2017sqlnet,yu2018naacl}. In \cite{zhong2017seq2sql} SEQ2SQL model used a coarse-grained sketch: aggregation, SELECT column and \texttt{WHERE} clause; 
\cite{xu2017sqlnet} used a finer sketch to align to the syntactical structure of a SQL query with three specific slot-filling models: model\_COL, model\_AGG, and model\_OPval. 
In  TypeSQL \cite{yu2018naacl} it also adopted this sketch-based model structures. However, in \cite{dong2018acl} sketch was referred to as abstractions for meaning representation, leaving out low-level details. This meaning sketch was used as an input to the final decoding. 

One of the challenge for using neural seq2seq models is the need of large annotated question-query pairs. \cite{zhong2017seq2sql,cai2018ijcai} have automatically generated large datasets using templates and had humans paraphrased the questions into natural language questions. WikiSQL is by far the largest text-to-SQL dataset. 
WikiSQL was designed for testing model's generalization by splitting the tables in a way that there is no overlap of tables in training and testing. 

Execution guided (EG) decoding was recently proposed in \cite{wang2018EG} that detects and excludes faulty outputs during the decoding by conditioning on the execution of partially generated output. Adding execution guided decoding to the coarse-to-fine model improved accuracy by $5.4\%$ on the wikiSQL dataset; and adding on top of the most recent IncSQL model \cite{Shi2018IncSQL} improved accuracy by $3.4\%$. It is proven that the EG module is very effective with any generative model.




Zero-shot semantic parsing has not obtained enough attention. \citet{herzig2018decoupling} applied a pipeline framework, including four independent models to achieve generalization, while our work is end-to-end trained and focusing on improving model's generalization with an auxiliary mapping task. Zero-shot slot filling \cite{bapna2017towards} also leverages the text of schema to connect language question words to column names (slots), but their model needs to predict the probability of each possible column indepentently while our model can select the column by processing the question and schema one time.

\section{Conclusions and Future Work}

In this paper, we propose a novel auxiliary mapping task for zero-shot text-to-SQL learning. Traditional seq2seq generation model is augmented with an explicit mapping model from question words to table schema. The generation model is first improved by an attentive pooling inside the question, and bi-directional attention flow to improve the interaction between the question and table schema. The mapping model serves as an enhancement model to text-to-SQL task as well as regularization to the generation model to increase its generalization. 

We compare our model with the a strong baseline coarse-to-fine model on the original WikiSQL testset as well as on the totally unseen test tables (a subset of zero-shot testing). Experimental results show that our model outperforms baseline models on both setting. Even though the generation model is already augmented with bi-directional attention to enhance the interaction between question and table, our results and analysis demonstrate that the explicitly mapping task can further increase the capability of generalization to unseen tables.


Spider \cite{Yu2018emnlp} was recently proposed as another large cross-domain text-to-SQL dataset besides WikiSQL.
It has more complex SQL templates including joint tables, which brings other interesting problems except for generalization. We plan to expand our models on this new dataset in the future.


\clearpage

\bibliography{emnlp-ijcnlp-2019}
\bibliographystyle{acl_natbib}

\clearpage
\appendix
\section{Appendices}
\label{sec:appendix}
\paragraph{Error case (a): wrong \texttt{COND\_COL} prediction}
{\small 
	\begin{itemize}
	\item Table: 2-11568882-2, Header: [year , winners , score , runners up , venue , 3rd place] \vspace{-0.2cm}
	\item Question: what 's in third place that 's going 1-0 ?\vspace{-0.2cm}
	\item Ground Truth: SELECT 3rd place FROM  2-11568882-2 WHERE  score = 1-0  \vspace{-0.2cm}
	\item Full-model Prediction: SELECT 3rd place FROM  2-11568882-2 WHERE  score = 1-0 \vspace{-0.2cm}
	\item Gen-model Prediction: SELECT 3rd place FROM  2-11568882-2 WHERE 3rd = 1-0 \vspace{-0.2cm}
	
    \end{itemize}
}

\paragraph{Error case (b): wrong \texttt{COND\_VAL} prediction}
{\small 
	\begin{itemize}
	\item Table: 1-1081235-1, Header: [name of lava dome , country , volcanic area , composition , last eruption or growth episode] \vspace{-0.2cm}
	\item Question: what countries have had eruptions of growth episodes in 1986 ? \vspace{-0.2cm}
	
	\item Ground Truth: SELECT country FROM  1-1081235-1 WHERE  last eruption or growth episode = 1986 \vspace{-0.2cm}
	\item Full-model Prediction: SELECT country FROM  1-1081235-1 WHERE  last eruption or growth episode = 1986 \vspace{-0.2cm}
	\item Gen-model Prediction: SELECT country FROM  1-1081235-1 WHERE  last eruption or growth episode = growth episodes in 1986 \vspace{-0.2cm}
    \end{itemize}
}

\paragraph{Error case (c): predicting extra conditions  or missing conditions}
{\small 
	\begin{itemize}
	\item Table: 2-11480171-1, Header: [year , title , genre , role , director] \vspace{-0.2cm}
	\item Question: what drama role does she play in 1973 ? \vspace{-0.2cm}
	\item Ground Truth: SELECT role FROM  2-11480171-1 WHERE  genre = drama AND year = 1973\vspace{-0.2cm}
	\item Full-model Prediction: SELECT role FROM  2-11480171-1 WHERE  genre = drama AND year = 1973 \vspace{-0.2cm}
	\item Gen-model Prediction:  SELECT role FROM  2-11480171-1 WHERE year = 1973 \vspace{-0.2cm}
	
    \end{itemize}
}

\end{document}